\documentclass[11pt,a4paper,pdflatex]{article}
\usepackage[hyperref]{acl2020}
\usepackage{times}
\usepackage{latexsym}
\usepackage{cite}
\usepackage{bm}
\usepackage{graphicx}
\usepackage{multirow}
\usepackage{threeparttable}
\usepackage{amsmath}
\usepackage{here}
\usepackage{CJKutf8}

\usepackage{microtype}

\aclfinalcopy 
\def\aclpaperid{7} 

\title{Memory Attentive Fusion: External Language Model Integration \\for Transformer-based Sequence-to-Sequence Model}

\author{Mana Ihori, Ryo Masumura, Naoki Makishima, \\ {\bf Tomohiro Tanaka, Akihiko Takashima, Shota Orihashi} \\
  NTT Media Intelligence Laboratories, NTT Corporation \\
  1-1 Hikarinooka, Yokosuka-Shi, Kanagawa 239-0847, Japan \\
  \texttt{mana.ihori.kx@hco.ntt.co.jp}}

\date{}

\begin{document}
\maketitle
\begin{abstract}
  This paper presents a novel fusion method for integrating an external language model (LM) into the Transformer based sequence-to-sequence (seq2seq) model.
  While paired data are basically required to train the seq2seq model, the external LM can be trained with only unpaired data.
  Thus, it is important to leverage memorized knowledge in the external LM for building the seq2seq model, since it is hard to prepare a large amount of paired data.
  However, the existing fusion methods assume that the LM is integrated with recurrent neural network-based seq2seq models instead of the Transformer.
  Therefore, this paper proposes a fusion method that can explicitly utilize network structures in the Transformer.
  The proposed method, called {\bf memory attentive fusion}, leverages the Transformer-style attention mechanism that repeats source-target attention in a multi-hop manner for reading the memorized knowledge in the LM.
  Our experiments on two text-style conversion tasks demonstrate that the proposed method performs better than conventional fusion methods.
\end{abstract}

\section{Introduction}
In recent studies, the Transformer sequence-to-sequence (seq2seq) model \citep{vaswani2017attention} has successfully performed in various natural language generation tasks, such as machine translation \citep{wang-etal-2019-learning,barrault-etal-2019-findings}, image captioning \citep{8682586,8869845,li2019entangled}, and automatic speech recognition \citep{dong2018speech,karita2019comparative,salazar2019self}. 
Although the Transformer training needs paired data, a large amount of paired data cannot often be prepared.
Moreover, unpaired data cannot be used for training the Transformer even though such data can be collected on a large scale.

To utilize a large amount of unpaired data, methods of integrating an external language model (LM) trained with these data into seq2seq models have been proposed \citep{shallow,gulcehre2015using,cold}.
These methods can improve the fluency of sentences that are generated by seq2seq models;
however, they integrate the LM into recurrent neural network (RNN) based seq2seq models rather than the Transformer.
In other words, LM fusion methods specific to the Transformer have not been considered yet.

Here, the Transformer employs the multi-hop attention mechanism \citep{sukhbaatar2015end} that repeats the source-target attention mechanism in each Transformer decoder block.
Thus, it is supposed that the source-target attention mechanism promotes to extract effective source information for target tasks more exactly than RNN based seq2seq models.
Therefore, we assume that the Transformer can utilize memorized knowledge in the external LM more effectively by using the multi-hop attention mechanism for the LM fusion.

In this paper, we propose a novel fusion method, called {\bf memory attentive fusion}, to integrate an external LM into the Transformer.
This fusion method utilizes a multi-hop source-target attention mechanism for combining the Transformer decoder with the external LM.
We performed experiments with two text-style conversion tasks: spoken-to-written style conversion and dialect conversion.
Our experiments demonstrate that the proposed method performs better than conventional fusion methods.

\section{Related work}
\begin{figure}[t]
  \centering
  \centerline{\includegraphics[clip, width=7.0cm]{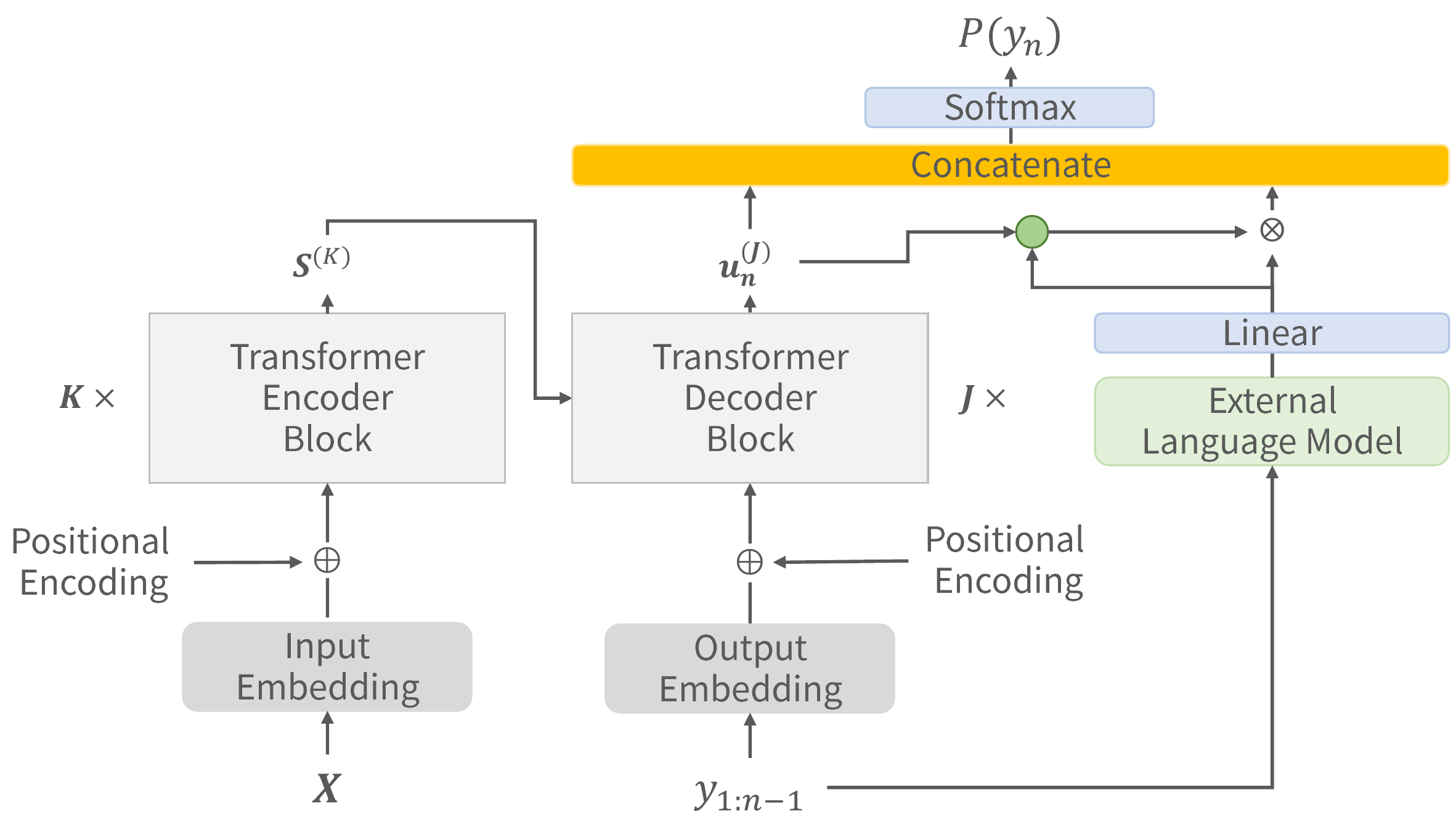}}
  \caption{Transformer with cold fusion.}
  \label{fig:cold}
\end{figure}

The simplest fusion method is to train the seq2seq model and the LM independently and then combine their outputs \citep{shallow,Chorowski2017TowardsBD,sutskever2014sequence}.
These methods are called shallow fusion.
Moreover, methods that integrate an external LM into seq2seq models during training have been proposed: deep fusion \citep{gulcehre2015using} and cold fusion \citep{cold}.
These methods utilize the information of not only paired data but also unpaired data in training.
Figure \ref{fig:cold} shows a Transformer with cold fusion.
These methods assume that the LM is integrated into RNN-based seq2seq models instead of the Transformer.

\section{Memory attentive fusion}
This section details memory attentive fusion for the Transformer seq2seq model.
In fact, memory attentive fusion is an extended method of the cold fusion \citep{cold}.
While the cold fusion uses memorized knowledge in the LM at an output layer only once, the memory attentive fusion repeatedly uses the knowledge at Transformer decoder blocks based on a source-target attention mechanism.

\begin{figure}[t]
  \centering
  \centerline{\includegraphics[clip, width=8.0cm]{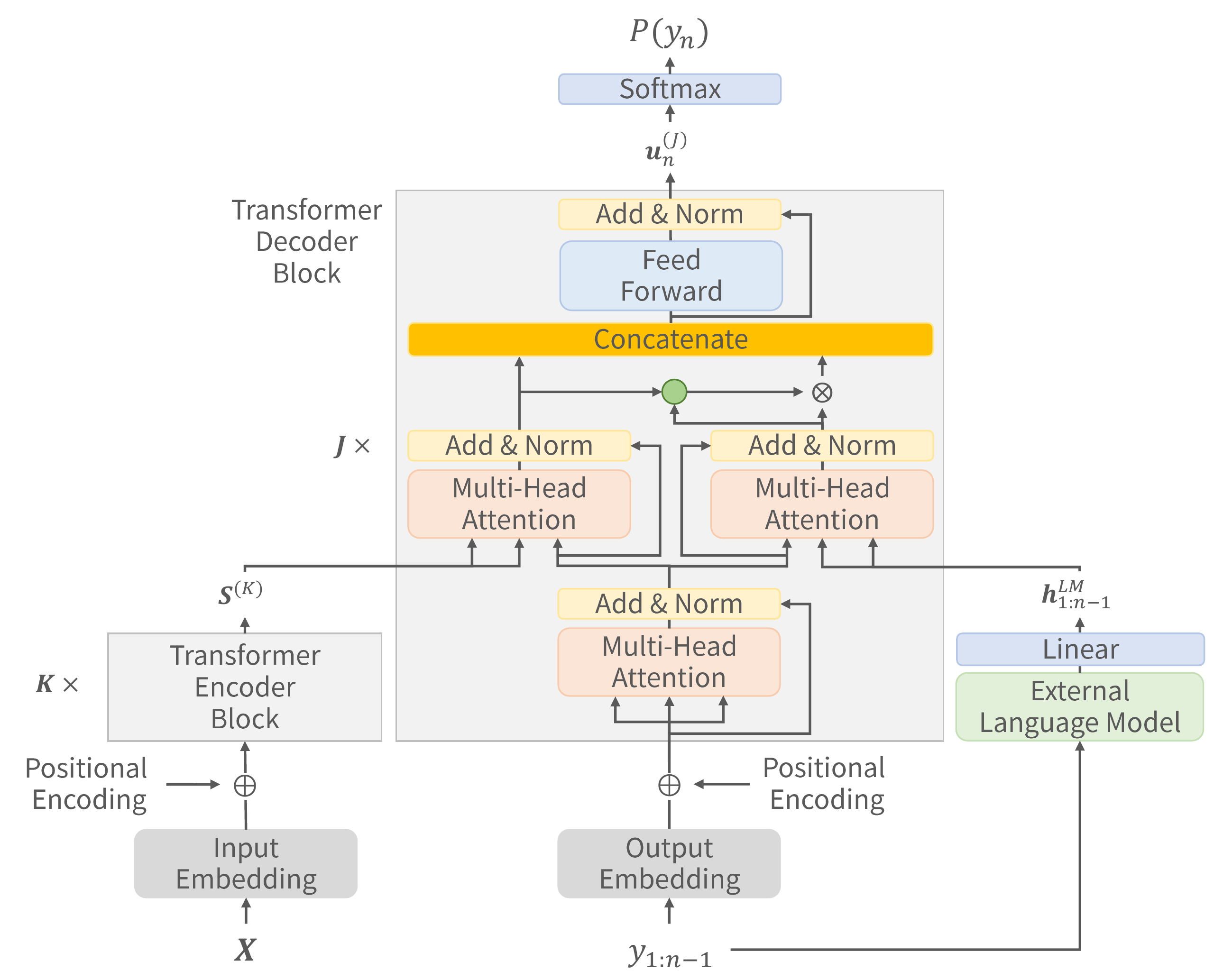}}
  \caption{Transformer with memory attentive fusion.}
  \label{fig:proposed}
\end{figure}

We define an input sequence as $\bm{X} = \{x_1, \cdots, x_M\}$ and an output sequence as $\bm{Y} = \{y_1, \cdots, y_N\}$, where $x_m$ and $y_n$ are tokens in the input and output sequence.
In text-style conversion, the model predicts the generation probabilities of the output sequence given the input sequence.
The generation probability of $\bm{Y}$ is defined as 
\begin{equation}
  P(\bm{Y}|\bm{X};\bm{\Theta}) = \prod_{n=1}^N P(y_n|y_{1:n-1}, \bm{X}; \bm{\Theta}) ,
\end{equation}
where $\bm{\Theta}=\{\theta_{\rm{enc}}, \theta_{\rm{dec}}, \theta_{\rm{lm}}\}$ represents model parameter sets.
$\theta_{\rm{enc}}$ and $\theta_{\rm{dec}}$ are trainable parameter sets with encoder and decoder, respectively.
$\theta_{\rm{lm}}$ is parameter set for the external LM.
$P(y_n|y_{1:n-1},$ $\bm{X}; \bm{\Theta})$ can be computed using an encoder and a decoder with memory attentive fusion in the Transformer.
Figure \ref{fig:proposed} shows the Transformer with memory attentive fusion.

\paragraph*{Encoder:}
The encoder converts an input sequence $\bm{X}$ into the hidden representations $\bm{S}^{(K)}$ using $K$ Transformer encoder blocks.
First, the input hidden representation of the Transformer encoder block $\bm{S}^{(0)} = \{\bm{s}_{1:M}^{(0)}\}$ is produced by 
\begin{equation}
  \bm{s}_m^{(0)} = {\tt Embedding}(x_m; \theta_{\rm{enc}}), 
\end{equation}
where ${\tt Embedding}(\cdot)$ consists of positional encoding and a linear layer.
Next, the $k$-th Transformer encoder block composes the $k$-th hidden representations $\bm{S}^{(k)}$ from the lower inputs $\bm{S}^{(k-1)}$ as
\begin{equation}
  \bm{S}^{(k)} = {\tt TransformerEncBlock}(\bm{S}^{(k-1)}; \theta_{\rm{enc}}),
\end{equation}
where ${\tt TransformeEncBlock}(\cdot)$ is the Transformer encoder block that consists of a scaled dot product multi-head self-attention layer and a position-wise feed-forward network \citep{vaswani2017attention}.

\paragraph*{Decoder with memory attentive fusion:}
The decoder with memory attentive fusion computes the generation probability of a token from the preceding tokens and hidden representations of the input sequence and the LM information.
The predicted probabilities of the $n$-th token $y_n$ are calculated as 
\begin{equation}
  P(y_n|y_{1:n-1}, \bm{X}) = {\tt softmax}(\bm{u}_{n}^{(J)}; \theta_{\rm{dec}}),
\end{equation}
where ${\tt softmax}(\cdot)$ is a softmax layer with a linear transformation.
The input hidden vector $\bm{u}_n^{(J)}$ is computed from $\bm{S}^{(K)}$ and $y_{1:n-1}$ using $J$ Transformer decoder blocks with an external LM.
First, the input hidden representation of the Transformer decoder block $\bm{u}_{n-1}^{(0)}$ and $\bm{h}_{n-1}^{\rm{LM}}$ are produced by 
\begin{align}
  \bm{u}_{n-1}^{(0)} &= {\tt Embedding}(y_{n-1}; \theta_{\rm{dec}}), \\
  \bm{l}_{n-1}^{\rm{LM}} &= {\tt LanguageModel}(y_{1:n-1}; \theta_{\rm{lm}}), \\
  \bm{h}_{n-1}^{\rm{LM}} &= {\tt linear}(\bm{l}_{n-1}^{\rm{LM}}; \theta_{\rm{dec}}), 
\end{align}
where ${\tt LanguageModel}(\cdot)$ is the trained external LM, and $\bm{l}_{n-1}^{\rm{LM}}$ is the logit output.
Next, we convert hidden representations in the lower layer $\bm{u}_{1:n-1}^{(j-1)}$ and the encoder output $\bm{S}^{(k)}$ into a hidden vector $\bm{c}_{n}^{(j)}$.
The hidden vector is computed as 
\begin{align}
  \bm{\bar{v}}_{n}^{(j)} &= {\tt SourceTarget}(\bm{u}_{1:n-1}^{(j-1)}, \bm{u}_{n-1}^{(j-1)}; \theta_{\rm{dec}}), \\
  \bm{v}_{n}^{(j)} &= {\tt LayerNorm}(\bm{u}_{n-1}^{(j-1)} + \bm{\bar{v}}_{n}^{(j)}), \\
  \bm{\bar{c}}_{n}^{(j)} &= {\tt SourceTarget}(\bm{S}^{(K)}, \bm{v}_{n}^{(j)}; \theta_{\rm{dec}}), \\
  \bm{c}_{n}^{(j)} &= {\tt LayerNorm}(\bm{v}_{n}^{(j)} + \bm{\bar{c}}_{n}^{(j)}),
\end{align}
where ${\tt SourceTarget}(\cdot)$ is a scaled dot product multi-head source target attention layer and ${\tt LayerNorm}(\cdot)$ is layer normalization \citep{ba2016layer}.
In memory attentive fusion, we also convert the LM output $\bm{h}_{1:n-1}^{\rm{LM}}$ and the hidden vector $\bm{v}_{n}^{(j)}$ into a hidden vector $\bm{b}_{n}^{(j)}$ with attention mechanism.
The hidden vector is computed as
\begin{align}
  \bm{\bar{b}}_{n}^{(j)} &= {\tt SourceTarget}(\bm{h}_{1:n-1}^{\rm{LM}}, \bm{v}_{n}^{(j)}; \theta_{\rm{dec}}), \\
  \bm{b}_{n}^{(j)} &= {\tt LayerNorm}(\bm{v}_{n}^{(j)} + \bm{\bar{b}}_n^{(j)}).
\end{align}
This attention mechanism is repeated with Transformer decoder block in the multi-hop manner.
Therefore, we expect to read the memorized memory in the LM effectively.
Next, we concatenate the hidden vector that have target and source information, and that have target and the LM information with gating mechanism by 

\begin{align}
  \bm{g}_{n}^{(j)} &= {\tt sigmoid}([{\bm{c}_{n}^{(j)}}^{\mathrm{T}}, {\bm{b}_{n}^{(j)}}^{\mathrm{T}}]^{\mathrm{T}}; \theta_{\rm{dec}}), \\
  \bm{q}_{n}^{(j)} &= [{\bm{c}_{n}^{(j)}}^{\mathrm{T}}, {\bm{g}_{n}^{(j)} \odot \bm{b}_{n}^{(j)}}^{\mathrm{T}}]^{\mathrm{T}}, 
\end{align}
where ${\tt sigmoid}(\cdot)$ is a sigmoid layer with a linear transformation.
Next, the hidden vector $\bm{q}_{n}^{(j)}$ is converted into the $j$-th hidden representation $\bm{u}_{n}^{(j)}$.
The hidden representation is computed as
\begin{align}
  \bm{\bar{u}}_{n}^{(j)} = {\tt FeedForward}(\bm{q}_{n}^{(j)}; \theta_{\rm{dec}}), \\
  \bm{u}_{n}^{(j)} = {\tt LayerNorm}(\bm{q}_{n}^{(j)} + \bm{\bar{u}}_{n}^{(j)}),
\end{align}
where ${\tt FeedForwrd}(\cdot)$ is a position-wise feed-forward network.

\paragraph{Training:}
In the Transformer, the model parameter set can be optimized from training dataset ${\cal D} = \{(\bm{X}^1, \bm{Y}^1), \cdots, (\bm{X}^{|\cal D|}, \bm{Y}^{|\cal D|})\}$.
The objective function for optimizing the model parameter is defined as
\begin{equation}
  {\cal L} = - \frac{1}{|{\cal D}|} \sum_{d=1}^{|{\cal D}|} \log P(\bm{Y}^d|\bm{X}^d; \bm{\Theta}). 
\end{equation}
Note that the external LM uses the freezing parameter $\theta_{\rm lm}$.

\section{Experiments}
We evaluated our method on text-style conversion tasks.
In particular, we chose {\it spoken-to-written style conversion task} and {\it dialect conversion task} in Japanese.
In the spoken-to-written style conversion task, spoken-style text produced by an automatic speech recognition system is converted into written-style text that has correct punctuation and no disfluency \citep{ihori2020parallel}.
In the dialect conversion task, Japanese dialects are converted into standard Japanese.

\subsection{Datasets}
\paragraph*{Spoken-to-written style conversion:}
We used the Corpus of Spontaneous Japanese (CSJ) \citep{maekawa2000spontaneous} and the parallel corpus for Japanese spoken-to-written style conversion (CJSW) \citep{ihori2020parallel}.
We divided the CSJ into a training set, validation set, and test set.
The training set, validation set, and test set have 46,847, 13,510, and 3,949 sentences, respectively.
The CJSW has four domains, and we divided it up following \citep{ihori2020parallel}.
We used all of the training and validation sets for training and each test set (CJSW-1, 2, 3, 4) for the evaluation.
All of these datasets are paired data of spoken-style text (manual transcriptions of speech) and written-style text (created with crowd-sourcing).

\paragraph*{Dialect conversion:}
We prepared three paired data of dialects (Tohoku-ben, Osaka-ben, Kyushu-ben) and standard Japanese with crowd-sourcing.
We divided these data into a training set, validation set, and test set for each dialect.
We used all of the training and validation sets for training and three test sets for the evaluation.
The training set, validation set and test set have 15,506, 3,924 and 2,160 sentences, respectively.
Moreover, the test set consists of 700 Tohoku-ben, 862 Osaka-ben, and 598 Kyushu-ben sentences, respectively.

\paragraph*{External text:}
We prepared a large-scale Japanese web text as the unpaired written-style text data.
The Web text was downloaded from various topic Web pages using our home-made crawler.
The downloaded pages were filtered in such a way that HTML tags, Javascript codes and other parts that were not useful for these tasks were excluded.
Finally, we prepare {\it one million sentences} for training the external LM.
Moreover, we divided this data into a training set, validation set.
The training set and validation set have 800,000 and 200,000 sentences, respectively.

\subsection{Setups}
\paragraph*{Transformer:}
We constructed the Transformer with shallow fusion \citep{shallow}, cold fusion \citep{cold} and memory attentive fusion methods.
In addition, we constructed the Transformer without fusion methods as a baseline.
We used the following configurations.
The dimensions of the output continuous representations and the inner outputs in the position-wise feed-forward network were set to 256, and the number of heads in the multi-head attentions was set to 8.
ReLU activation was used in nonlinear transformation function.
We stacked 4-layer Transformer encoder blocks, and 2-layer Transformer decoder blocks.
We set the output unit size (witch corresponded to the amount of tokens in the training set for language model) to 5,640.
To train these models, we used the adam optimizer and label smoothing with a smoothing parameter of 0.1.
The training steps were stopped based on early stopping using the part of the training data.
We set the mini-batch size to 64 sentences and the dropout rate in the Transformer blocks to 0.2.
For the mini-batch training, we truncated each sentence to 200 tokens.
We used characters as tokens.
All trainable parameters were randomly initialized.
For the decoding, we used a beam search algorithm in which the beam size was set to 4.

\paragraph*{External LM:}
We utilized Open AI GPT \citep{radford2019language} for the LM fusion, although any LM can potentially be used.
We used the following configurations.
The number of heads in the multi-head attentions was set to 4.
We stacked 8-layer Transformer blocks.
The training steps were stopped based on early stopping using the part of the training data.
We set the dropout rate in the Transformer blocks to 0.1.
The other settings were the same as the Transformer settings.
After training, perplexity of this LM was 11.8.
Note that this LM was used in both two tasks and the Transformer and the external LM were not pre-trained.

\subsection{Results}
Table \ref{table:csj} shows the experimental results in the spoken-to-written style conversion task.
Also, Table \ref{table:dialect} shows the experimental results in the dialect conversion task.
We calculated automatic evaluation scores in three metrics: BLEU-3 (B-3) \citep{papineni2002bleu}, ROUGE-L (R-L) \citep{lin2004automatic}, and METEOR \citep{banerjee-lavie-2005-meteor}.
{\it Baseline} in the tables mean the results of the Transformer without the external LM. 

Table \ref{table:csj} shows that shallow fusion and cold fusion performed worse than the baseline on the CSJ dataset.
On the other hand, memory attentive fusion outperformed the baseline.
Moreover, memory attentive fusion outperformed the baseline and shallow fusion on the CJSW dataset.
In addition, cold fusion outperformed the baseline on CJSW-1, -3 and -4.
As in the spoken-to-written style conversion task, Table \ref{table:dialect} shows that memory attentive fusion outperformed the other methods. 

The above results show that shallow fusion is not suitable for the Transformer because it degraded performance in all cases.
Moreover, when the LM was integrated with cold fusion, the performance was better than baseline in some domains.
Thus, we consider that cold fusion can be used with the Transformer in limited cases.
\begin{table}[H]
  \centering
  \small
  \begin{threeparttable}
  \begin{tabular}{|c|crrr|} 
      \hline
      Test set & & B-3 & R-L & METEOR \\
       \hline \hline
       \multirow{4}{*}{CSJ} & a). & 0.667 & 0.855 & 0.853 \\
       & b). & 0.667 & 0.850 & 0.853 \\
       & c). & 0.657 & 0.852 & 0.847 \\
       & d). & \bf{0.669} & \bf{0.860} & \bf{0.856} \\
       \hline
       \hline
       \multirow{4}{*}{CJSW-1} & a). & 0.723 & 0.785 & 0.881 \\
       & b). & 0.705 & 0.775 & 0.870 \\
       & c). & 0.734 & 0.791 & \bf{0.889} \\
       & d). & \bf{0.735} & \bf{0.792} & 0.887 \\
       \hline
       \multirow{4}{*}{CJSW-2} & a). & 0.657 & 0.718 & 0.847 \\
       & b). & 0.630 & 0.702 & 0.830 \\
       & c). & 0.655 & 0.714 & 0.840 \\
       & d). & \bf{0.671} & \bf{0.726} & \bf{0.859} \\
       \hline
       \multirow{4}{*}{CJSW-3} & a). & 0.671 & 0.732 & 0.839 \\
       & b). & 0.663 & 0.726 & 0.836 \\
       & c). & 0.672 & 0.73 & 0.842 \\
       & d). & \bf{0.686} & \bf{0.737} & \bf{0.85} \\
       \hline
       \multirow{4}{*}{CJSW-4} & a). & 0.772 & 0.818 & 0.898 \\
       & b). & 0.752 & 0.806 & 0.884 \\
       & c). & 0.775 & 0.819 & 0.897 \\
       & d). & \bf{0.779} & \bf{0.821} & \bf{0.900} \\
       \hline
  \end{tabular}
  \begin{tablenotes} \footnotesize
    \item a). Baseline \hspace{2mm} b). Shallow fusion
    \item c). Cold fusion \hspace{2mm} d). Memory attentive fusion
  \end{tablenotes} 
  \end{threeparttable}
  \caption{\label{table:csj} Results on spoken-to-written style conversion tasks.}
\end{table}
On the other hand, memory attentive fusion outperformed the other fusion methods in almost all of the domains and tasks.
Therefore, we consider that memory attentive fusion is suitable for integration of the external LM into the Transformer.
In addition, memory attentive fusion worked well especially in the dialect conversion task.
Thus, we can infer that the fusion method for the Transformer is more effective when there is small training data.

\begin{figure}[tb]
  \centering
  \centerline{\includegraphics[clip, width=7.0cm]{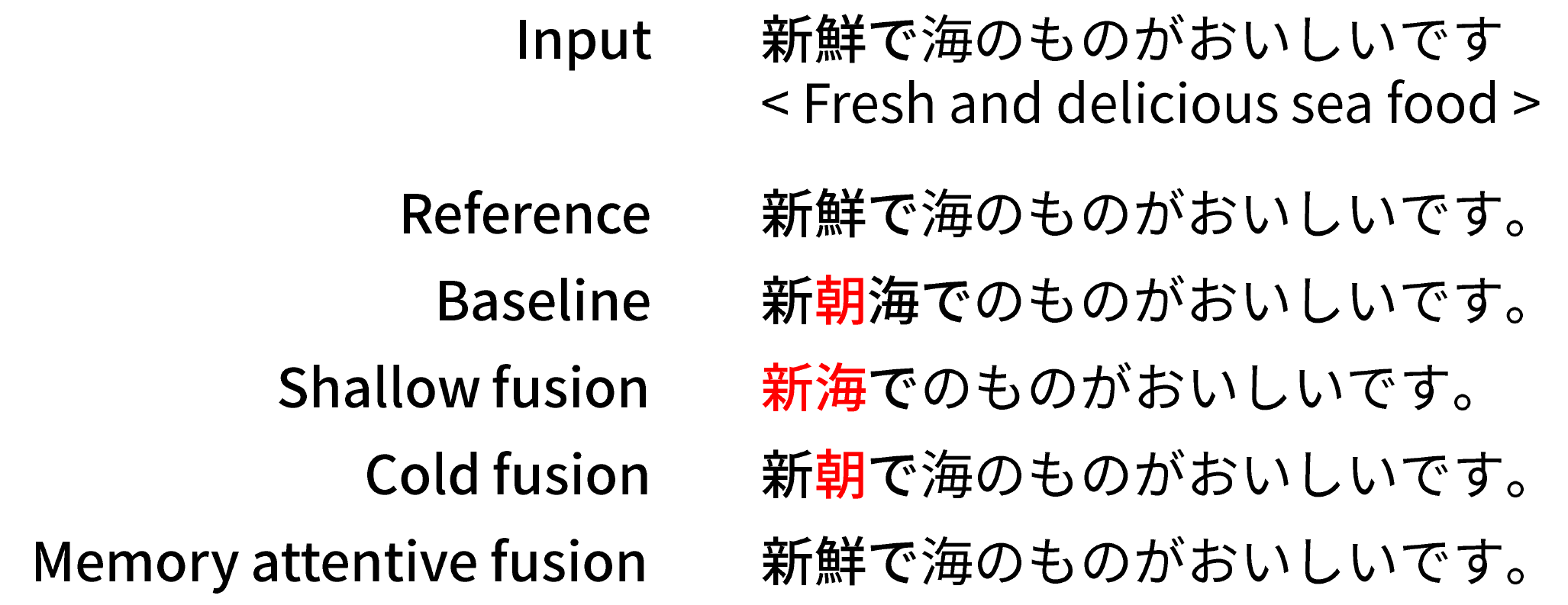}}
  \caption{Example of spoken-to-written style conversion in CSJ dataset with each fusion method.}
  \label{fig:example}
\end{figure}

We show the converted example of spoken-to-written style conversion in CSJ dataset with each fusion method in Figure \ref{fig:example}.
Figure \ref{fig:example} shows that the word ``\begin{CJK}{UTF8}{min}新鮮\end{CJK}'' (flesh) was output correctly with memory attentive fusion, but other methods were not output the word correctly.
The word ``\begin{CJK}{UTF8}{min}新鮮\end{CJK}'' was not included in training data for the Transformer, but it was included in training data for the external LM.
This indicate that only memory attentive fusion was successful in extracting knowledge of the external LM.

\begin{table}[tb]
  \centering
  \small
  \begin{threeparttable}
  \begin{tabular}{|c|crrr|} 
      \hline
       Test set & & B-3 & R-L & METEOR \\
       \hline \hline
       \multirow{4}{*}{Osaka-ben} & a). & 0.649 & 0.784 & 0.790 \\
       & b). & 0.638 & 0.774 & 0.780 \\
       & c). & 0.648 & 0.784 & 0.787 \\
       & d). & \bf{0.663} & \bf{0.795} & \bf{0.802} \\
       \hline 
       \multirow{4}{*}{Kyushu-ben} & a). & 0.741 & 0.857 & 0.872 \\
       & b). & 0.729 & 0.849 & 0.859 \\
       & c). & 0.738 & 0.855 & 0.867 \\
       & d). &\bf{0.752} & \bf{0.864} & \bf{0.880} \\
       \hline 
       \multirow{4}{*}{Tohoku-ben} & a). & 0.619 & 0.767 & 0.742 \\
       & b). & 0.603 & 0.755 & 0.721 \\
       & c). & 0.610 & 0.761 & 0.730 \\
       & d). & \bf{0.630} & \bf{0.772} & \bf{0.752} \\
       \hline
  \end{tabular}
  \begin{tablenotes} \footnotesize
    \item a). Baseline \hspace{2mm} b). Shallow fusion
    \item c). Cold fusion \hspace{2mm} d). Memory attentive fusion
  \end{tablenotes} 
  \end{threeparttable}
  \caption{\label{table:dialect} Results on dialect conversion tasks.}
\end{table}

\section{Conclusion}
We proposed memory attentive fusion, a novel method to integrate an external LM into the Transformer.
Conventional fusion methods assume that the LM is integrated into the RNN-based seq2seq. 
On the other hand, the proposed method employs a Transformer-specific fusion method which repeats the attention mechanism for the LM many times.
Experiments demonstrated that the proposed method outperformed the conventional methods in two tasks.
We conclude that the proposed method is suitable for integrating the LM into the Transformer.
In the future work, we will try using the proposed method in other natural language generation tasks such as automatic speech recognition.


\end{document}